\documentclass[letterpaper, 10 pt, conference]{ieeeconf}  

\IEEEoverridecommandlockouts                              

\usepackage{times} 
\usepackage{amsmath} 
\usepackage{amssymb}  

\usepackage{graphicx}
\graphicspath{ {images/} {data/} }

\usepackage{verbatim}
\usepackage{array}
\newcolumntype{C}[1]{>{\centering\arraybackslash}p{#1}}
\usepackage{bm}
\newcommand{\bsigma}{\bm{\sigma}}

\usepackage{mathtools}
\DeclarePairedDelimiter{\abs}{\lvert}{\rvert}
\usepackage{siunitx}
\usepackage{multirow}
\usepackage{tabularx}
\usepackage{makecell}
\usepackage{tablefootnote}
\usepackage{pifont}

\usepackage{xspace}
\makeatletter
\DeclareRobustCommand\onedot{\futurelet\@let@token\@onedot}
\def\@onedot{\ifx\@let@token.\else.\null\fi\xspace}
\def\eg{\emph{e.g}\onedot}

\def\ie{\emph{i.e}\onedot}

\makeatother

\newcommand*\rot{\rotatebox[origin=l]{90}}

\title{\LARGE \bf
On Boosting Semantic Street Scene Segmentation\\ with Weak Supervision
}

\author{Panagiotis Meletis$^{}$ and Gijs Dubbelman$^{}$
	\thanks{$^{}$Panagiotis Meletis ({\tt\small p.c.meletis@tue.nl}) and Gijs Dubbelman ({\tt\small g.dubbelman@tue.nl}) are with the Department of Electrical Engineering, Eindhoven University of Technology,
	Eindhoven,
	The Netherlands.
	This project has received funding from the European Union’s Horizon 2020 research and innovation programme under grant agreement No 688099.
    }%
}

\begin{document}

\maketitle
\thispagestyle{empty}
\pagestyle{empty}

\begin{abstract}
Training convolutional networks for semantic segmentation requires per-pixel ground truth labels, which are very time consuming and hence costly to obtain. Therefore, in this work, we research and develop a hierarchical deep network architecture and the corresponding loss for semantic segmentation that can be trained from weak supervision, such as bounding boxes or image level labels, as well as from strong per-pixel supervision. We demonstrate that the hierarchical structure and the simultaneous training on strong (per-pixel) and weak (bounding boxes) labels, even from separate datasets, consistently increases the performance against per-pixel only training. Moreover, we explore the more challenging case of adding weak image-level labels. We collect street scene images and weak labels from the immense Open Images dataset to generate the OpenScapes dataset, and we use this novel dataset to increase segmentation performance on two established per-pixel labeled datasets, Cityscapes and Vistas. We report performance gains up to +13.2\% mIoU on crucial street scene classes, and inference speed of 20 fps on a Titan V GPU for Cityscapes at 512 x 1024 resolution. Our network and OpenScapes dataset are shared with the research community.
\end{abstract}

\section{Introduction}
Semantic segmentation of street scene images is a fundamental building block for automated driving~\cite{janai17computer}. It is the first step of scene understanding and provides the necessary information towards higher level reasoning and planning~\cite{wilko2018planning}. Formulation of the problem as per-pixel (dense) classification and modeling it with Fully Convolutional Networks~\cite{long2015fully} has become the de facto solution for semantic segmentation of images. However, its success is based on the availability of huge amounts of tediously, per-pixel labeled datasets~\cite{Cordts2016Cityscapes, neuhold2017mapillary, huang2018apolloscape}, and existing solutions do not leverage weakly labeled data that are provided in larger and more diverse datasets~\cite{kuznetsova2018open}.

Therefore, in this work, we explore a method for per-pixel training of Fully Convolutional Networks on multiple datasets simultaneously, containing images with strong (per-pixel) or weak (bounding boxes and image-level) labels. The ability to train from weakly labeled data is an important research topic in the field of computer vision~\cite{papandreou2015weakly, kolesnikov2016seed, pinheiro2015image, ye2018diverse}, which, when solved, can be of benefit to many application domains.

The challenge when using weak supervision for per-pixel semantic segmentation lies on the different and incompatible \textit{annotation types}~\cite{meletis2018heterogeneous}.
Our method fully solves, in a consistent and uniform manner that challenge, while training on heterogeneous datasets. It consists of a hierarchy of classifiers and a hierarchical loss function, which can handle any of the aforementioned types of supervision. This is achieved by transforming the incompatible, for semantic segmentation, weak labels into per-pixel weak labels, a common practice also appearing in other related problems~\cite{lu2019occupancy}.


\begin{figure}
	\begin{center}
		\includegraphics[width=1.0\linewidth]{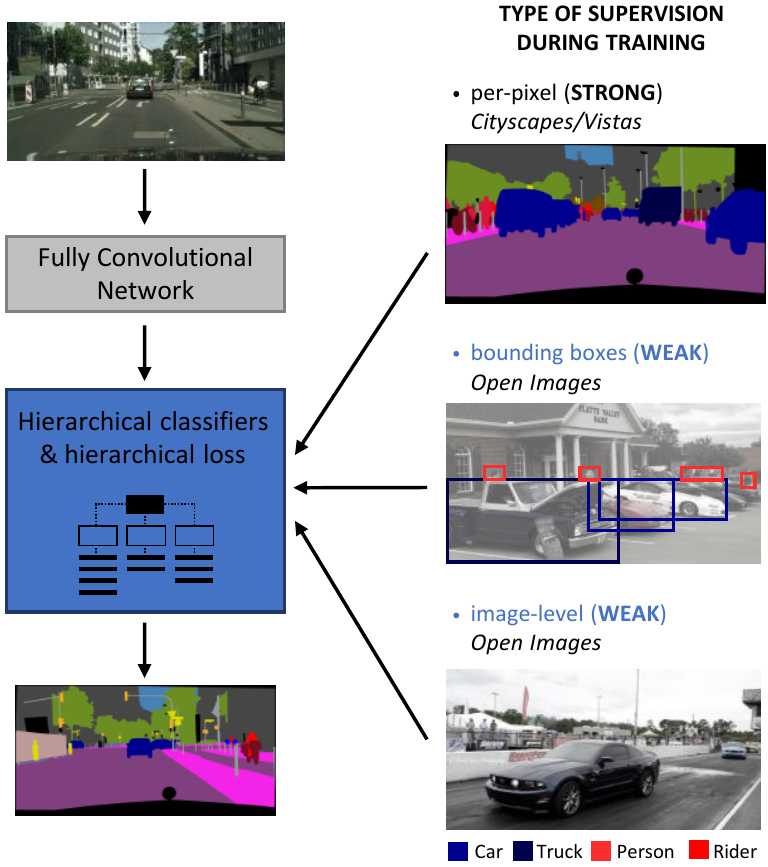}
	\end{center}
	\caption{Our contributions in blue color. Using diverse types of weak supervision from the Open Images dataset we achieve increase on performance over datasets with strong supervision using a fully convolutional network with a hierarchy of classifiers and the corresponding hierarchical loss.}
	\label{fig:eye-catcher}
\end{figure}

\begin{figure}
	\begin{center}
		\includegraphics[width=1.0\linewidth]{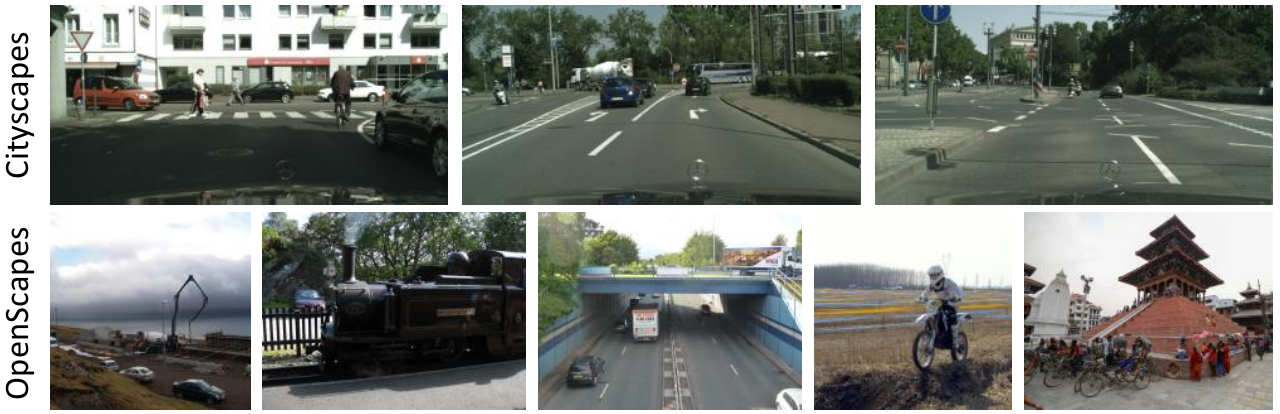}
	\end{center}
	\caption{Example images from per-pixel labeled Cityscapes dataset and the weakly labeled \textit{OpenScapes} dataset that demonstrate the big \textit{domain gap}.}
	\label{fig:examples-domain-gap}
\end{figure}

In order to prove how weak labels can boost semantic segmentation performance of strongly (per-pixel) labeled datasets, \eg Cityscapes~\cite{Cordts2016Cityscapes}, we collect a weakly labeled dataset by mining street scene images from the very large scale Open Images dataset~\cite{kuznetsova2018open} and we name it \textit{OpenScapes}. This new dataset contains 100,000 images with 2,242,203 bounding box labels and 100,000 images with 1,199,582 image-level labels, and spans 14 of the most important street scene classes. However, even after the automated selection procedure, the \textit{domain gap} between \textit{OpenScapes} and the per-pixel datasets, for which we want to prove the performance increase, remains large, as can be seen in Fig.~\ref{fig:examples-domain-gap}, making the weak supervision even "weaker".

We evaluate our system on two established per-pixel labeled datasets and we show that performance increase is proportional to the amount of extra weak labels used. We achieve that without using any external components as in \cite{ye2018diverse, papandreou2015weakly}, but only the hierarchical structure of the classifiers and the proposed hierarchical loss.


To summarize, the contributions of this work are:
\begin{itemize}
	\item A methodology for training semantic segmentation networks on datasets with diverse supervision, including per-pixel, bounding box, and image-level labels.
	\item \textit{OpenScapes} dataset: a large, weakly labeled dataset with $200,000$ images and 14 semantic classes for street scenes recognition.
\end{itemize}

Our system and the \textit{OpenScapes} dataset are made available to the research community~\cite{panos2019code}.

\begin{figure}
	\begin{center}
		\includegraphics[width=1.0\linewidth]{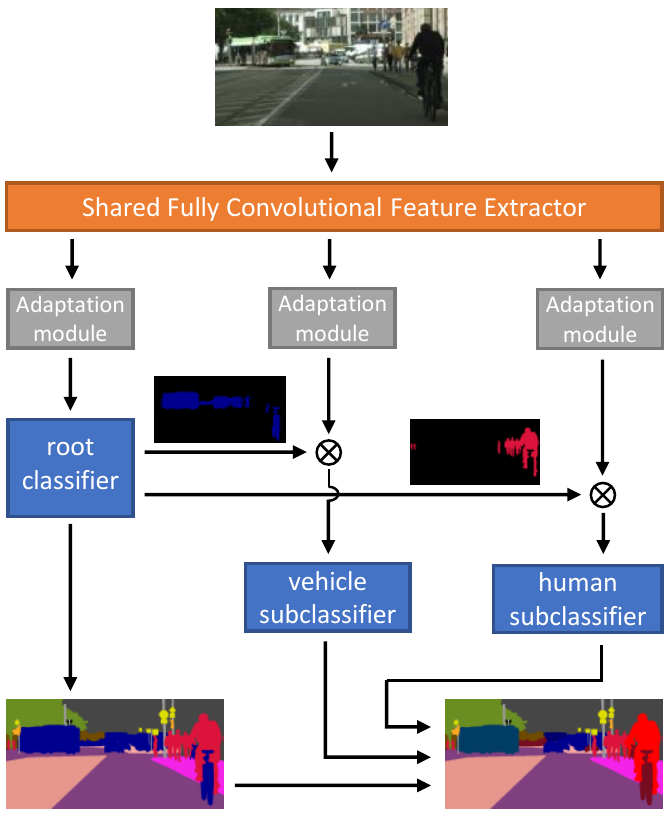}
	\end{center}
	\caption{Network architecture. The root classifier passes its decisions to the two subclassifiers, which classify only pixels that are assigned to them by the root classifier.}
	\label{fig:architecture}
\end{figure}

\section{Method}
In this Section, we describe the proposed training and inference methodology. We generalize previous work~\cite{meletis2018heterogeneous} by enabling weak supervision from both bounding box labeled and image-level labeled images without using any external components.

Our method facilitates training of any fully convolutional network for per-pixel semantic segmentation and only requires a specific structure of classifiers and a specialized loss to train them. To achieve that, weak labels (bounding boxes and image labels) have to be converted to pseudo per-pixel ground truth. This is described in Sec.~\ref{ssec:pp-gt-gen}. The network architecture and the corresponding hierarchical loss are presented in Sec.~\ref{ssec:conv-net-arch} and Sec.~\ref{ssec:hier-loss} respectively. In addition, we address the shortcomings of pseudo ground-truth generation~\cite{meletis2018heterogeneous} for any type of weak labels.

\subsection{Convolutional Network Architecture}
\label{ssec:conv-net-arch}


The network architecture follows the design proposed in~\cite{meletis2018heterogeneous} and is depicted in Fig.~\ref{fig:architecture}. Specifically, we opt for a two-level hierarchical convolutional network, which consists of a fully convolutional shared feature extractor and a set of, hierarchically arranged, classifiers. The root classifier is trained only with strong supervision (per-pixel labeled semantic classes). The subclassifiers are trained, using the hierarchical loss of Sec.~\ref{ssec:hier-loss} with per-pixel supervision. For that purpose, the weak labels are converted into per-pixel pseudo ground truth as described in Sec.~\ref{ssec:pp-gt-gen}.

The benefits of the hierarchical structure~\cite{meletis2018heterogeneous} are twofold: 1) it solves the problem of simultaneously training with different types of supervision, by placing classes with weak labels in the subclassifiers, and 2) it solves the semantic class incompatibilities between datasets, due to the unavailability of specific semantic classes in all datasets.

The hierarchy of classifiers is constructed according to the availability of strong and weak labels for each class. The root classifier (left in Fig.~\ref{fig:architecture}) contains high-level classes with per-pixel labels. Each one of the subclassifiers corresponds to one high-level class of the root classifier, and contains subclasses with per-pixel and/or weak supervision.

The shared feature representation (see Fig.~\ref{fig:architecture}) is passed through two shallow, per-classifier \textit{adaptation networks}, which adapt the common representation, its depth, and receptive field to meet the requirements of each classifier as described in~\cite{meletis2018heterogeneous}. In this work, we use a single ResNet bottleneck layer~\cite{he2016deep} as in~\cite{meletis2018heterogeneous}.


\subsection{Generation of pseudo per-pixel ground truth from weak labels}
\label{ssec:pp-gt-gen}

The goal is to train the network with per-pixel labels, thus we need to generate per-pixel ground truth from bounding boxes and image-level labels. The 2D ground truth generation procedure in~\cite{meletis2018heterogeneous} is ambiguous for classes whose bounding box boundaries do not match tightly the object boundaries. Thus that method is valid only for square-shaped, compact objects, like traffic signs, and cannot be applied to image-level labels.

\begin{figure}
	\centering
	\includegraphics[width=1.0\linewidth]{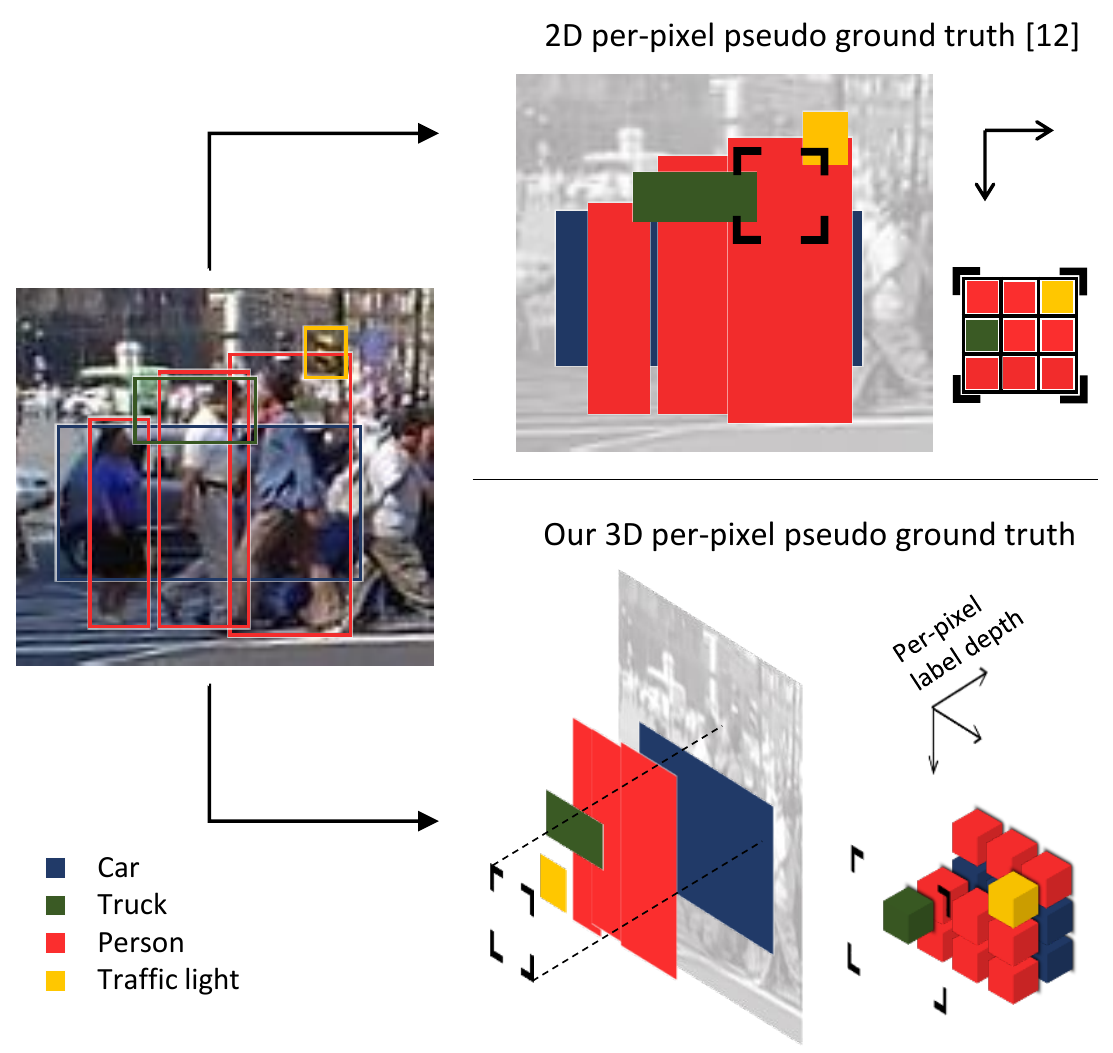}
	\caption{2D vs 3D per-pixel pseudo ground truth (GT) generation. Left: image with a selected subset of bounding boxes colored by class. Right top: 2D GT generation used in~\cite{meletis2018heterogeneous}. Right bottom: proposed 3D GT generation. In 2D GT, overlapping bounding boxes produce ambiguity in generated per-pixel labels (\eg the car label is hidden behind pedestrian labels), which is solved by adding a 3rd dimension in GT generation. The pseudo ground truth has the form of a dense categorical instead of a sparse (one-hot) distribution. The same principle is used for generating GT from image level labels by considering for each label the boundaries of the bounding box to extend to the whole image.}
	\label{fig:2d-vs-3d-gt-gen}
\end{figure}

According to~\cite{meletis2018heterogeneous} the 2D pseudo ground truth for each image is generated pixel-wise, by assigning a single label to each pixel from the set of bounding boxes that this pixel belongs to, as depicted at the top of Fig.~\ref{fig:2d-vs-3d-gt-gen}. This procedure effectively generates a so-called sparse or one-hot categorical probability distribution, since each pixel belongs to a specific class with probability $1$. Contrary, in this work, we model the per-pixel labels as a dense or multi-hot categorical probability distribution, and thus the ground truth for each images becomes 3D (see Fig.~\ref{fig:2d-vs-3d-gt-gen}). This model assigns to each pixel a probability for every class, and the sum of probabilities for all classes must be $1$. In order to convert bounding boxes and image labels to per-pixel labels, we use a voting scheme, according to which each label increases each pixel's counter vector by $1$. After collecting all votes we normalize across all classes, in order for the labels to represent a valid probability distribution.

\subsection{Hierarchical loss}
\label{ssec:hier-loss}

\begin{table}
	\caption{Loss components per classifier and per dataset. All losses are per-pixel Categorical Cross Entropy (CCE) losses between the dense or sparse categorical labels and the softmax probabilities of the associated classifier.}
	\label{tab:loss-components}
	\begin{center}
		\begin{tabular}{c|c|c}
			\multirow{2}{*}{classifier} & Per-pixel labeled data & Weakly labeled data \\
			& (Cityscapes or Vistas) & (OpenScapes) \\
			\hline
			
			root & sparse CCE & - \\
			vehicle subcl. & dense CCE & conditional dense CCE \\
			human subcl. & dense CCE & conditional dense CCE
		\end{tabular}
	\end{center}
\end{table}

We construct the hierarchical loss similar to~\cite{meletis2018heterogeneous}, namely the loss is accumulated unconditionally for per-pixel labeled datasets and conditionally for per bounding box or per image-level labeled datasets. The total loss terms are pixel-wise categorical cross entropy losses and are summarized in Table~\ref{tab:loss-components}. The five loss terms of Table~\ref{tab:loss-components} are added, using coefficients of $0.1$ for the subclassifier's losses and, together with the regularization loss, make up the total loss.

We assign a set of $n$ classes named $\{1, ..., n\}$ for each classifier. The per-pixel labels are given in the form of a vector $\mathcal{Y} = [y_1, ..., y_n]$ with elements corresponding to per-class probabilities from a categorical distribution,~\ie $\sum_{y \ in ~\mathcal{Y}} y = 1$. The general form of the per-pixel categorical cross-entropy loss for a softmax classifier with $n$ classes and softmax output $\bsigma = [\sigma_{y_1}, ..., \sigma_{y_n}]$ for all pixels $\mathcal{P}$ is:

\begin{equation}
\label{eq:gen-loss}
\mathcal{L} = -\frac{1}{\abs{\mathcal{P}}} \sum_{p \in \mathcal{P}} \sum_{y \in \mathcal{Y}} y \log \sigma_y
\end{equation}

For the root classifier, we use only the sparse categorical per-pixel labels, since this classifier receives supervision from the per-pixel labeled dataset. In this case, $y_j = 1$ for the class $j$ that the pixel is labeled with, and $y_{i \neq j} = 0$ for all other classes. Thus, Eq.~\ref{eq:gen-loss} is reduced to sparse CCE loss:

\begin{equation}
\label{eq:sparse-cce}
\mathcal{L} = -\frac{1}{\abs{\mathcal{P}}} \sum_{p \in \mathcal{P}} \log \sigma_{y_j}
\end{equation}

For the subclassifiers, we use dense categorical per-pixel labels for both the per-pixel and the weakly labeled images (see Sec.~\ref{ssec:pp-gt-gen}). We convert the per-pixel labeled dataset's sparse labels to dense categorical labels by assigning a probability of $y_j = 1$ to the ground truth class $j$, and probability $y_{i \neq j} = 1$ to all other classes. For per-pixel labeled images we use Eq.~\ref{eq:sparse-cce}. For weakly labeled images we use the loss of Eq.~\ref{eq:gen-loss}, which is accumulated from pixels $\mathcal{P}$ that satisfy two conditions: 1) the per-pixel pseudo ground truth has non-zero probability for that pixel, and 2) the root classifier decision agrees with the per-pixel pseudo ground truth for that pixel,~\ie it is a class that has non-zero probability in the per-pixel pseudo ground truth.

\begin{table}
	\caption{OpenScapes dataset overview and comparison with per-pixel labeled datasets. Training splits are shown.}
	\label{tab:openscapes-vs-others}
	\begin{center}
		\begin{tabular}{c|c|c|c}
			& Cityscapes & Vistas & OpenScapes \\
			\hline
			
			\# of images & 2975 & 18000 & 200,000 \\
			\# of classes & 27 & 65 & 14 \\
			\# of pixel labels & $ ~1.6 \cdot 10^9 $ & $ ~156.2 \cdot 10^9 $ & - \\
			\# of bound. boxes & - & - & 2,242,203 \\
			\# of image labels & - & - & 1,199,582
		\end{tabular}
	\end{center}
\end{table}

\section{OpenScapes Dataset and Implementation}
In this Section, we describe the collection process of \textit{OpenScapes} and we compare it with the per-pixel annotated Cityscapes and Vistas datasets. Moreover, we discuss all the implementation details for our experiments.

\subsection{OpenScapes Street Scenes Dataset}
\label{ssec:openscapes}
We collect images of street scenes from the recently open-sourced, very large scale Open Images dataset~\cite{kuznetsova2018open} and create a subset that we call \textit{OpenScapes}. Open Images dataset contains over 9,000,000 images, 14,600,000 bounding boxes for 600 object classes, and more than 27,900,000 human-verified image-level labels for 19,794 classes. We collected 200,000 images, containing 2,242,203 bounding box labels and 1,199,582 image-level labels from 14 classes, with as much as possible street scene related content. 

The fully automated collection procedure is described in Sec.~\ref{ssec:mining-proc}. However, even after the careful selection, the \textit{domain gap}~\cite{zhang2017transfer, csurka2017domain} between the per-pixel datasets (Cityscapes, Vistas) and \textit{OpenScapes} is large. This can be seen by the image examples in Figures~\ref{fig:eye-catcher} and~\ref{fig:examples-domain-gap}, and is discussed together with a comparison with the employed per-pixel labeled datasets in Sec.~\ref{ssec:openscapes-vs-others}.

\subsubsection{\textbf{Mining procedure}}
\label{ssec:mining-proc}
First, we rank in descending order images from Open Images by the number of bounding boxes and image-level labels they contain for the 14 selected street scene classes. Then we select the top 100,000 images for the bounding box labeled subset and then 100,000 images for the image-level labeled subset and we make sure that there is no image overlap between the two subsets. For the ranking we used a voting system, according to which classes in the weak labels of an image vote for an image to be a street scene image or not. The more probable classes, like traffic light and license plate, can cast more votes than classes that may appear in other contexts (\eg car, person). 

\subsubsection{\textbf{Comparison with per-pixel labeled datasets}}
\label{ssec:openscapes-vs-others}
In Tab.~\ref{tab:openscapes-vs-others} we compare \textit{OpenScapes} with two established per-pixel labeled datasets that we experiment on also in this paper. In Fig.~\ref{fig:examples-domain-gap} we present some images from Cityscapes and \textit{OpenScapes}. As can be seen Cityscapes image domain is very consistent with images taken from a specific point of view and in one country, contrary to \textit{OpenScapes}, which contains web-like images and does not correspond to a consistent domain.

\begin{figure}
	\centering
	\includegraphics[width=1.0\linewidth]{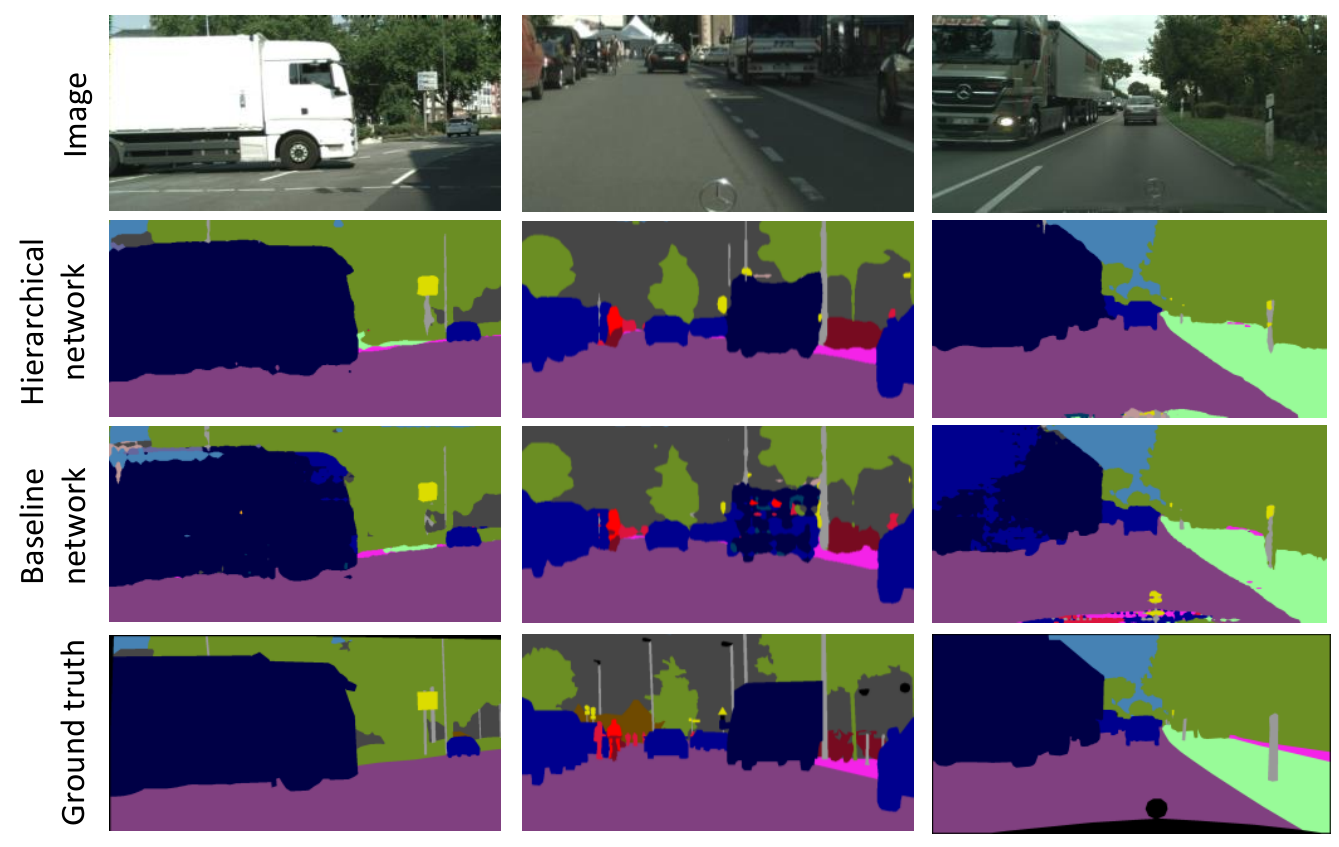}
	\caption{Comparison for Cityscapes validation split images for the hierarchical network trained on \textit{OpenScapes} and Cityscapes against the baseline network trained on Cityscapes only. The three classes with biggest improvement in mIoU are Truck(+13.2\%), Rider(+3.5\%), and Person(+2.1\%).}
	\label{fig:citys-res}
\end{figure}

\subsection{Implementation details}
\label{ssec:impl-details}
The network is depicted in Fig.~\ref{fig:architecture}. The feature extractor consists of the ResNet-50 layers (without the classifier) from~\cite{he2016deep}, followed by an 1x1 convolutional layer, to decrease feature dimensions to 256, and a Pyramid Pooling Module~\cite{zhao2017pspnet}. The stride of the feature representation on the input is reduced from 32 to 8, using dilated convolutions. Each branch has an extra bottleneck module~\cite{he2016deep}, a bilinear upsampling layer to recover original resolution, and a softmax classifier.

We use Tensorflow~\cite{abadi2016tensorflow} and 4 Titan V 12 GB GPUs for training. We implemented synchronous, cross-GPU batch normalization, and for all experiments we use batch size of up to 4 images per-GPU depending on the experiment, containing 1 image from the per-pixel labeled dataset (Cityscapes or Vistas), 2 images from the bounding box labeled dataset (\textit{OpenScapes} subset), and 1 image from the image-level labeled dataset (\textit{OpenScapes} subset).

For experiments involving Cityscapes we use images dimensions of 512x1024 and for Vistas 621x855. Since, \textit{OpenScapes} images have multiple aspect ratios, we upscale each image to fit tightly the aspect ratio of the per-pixel labeled dataset and then we crop a random patch of same dimensions as the per-pixel labeled image. Networks with batch size of 3 per-GPU are trained for 26 epochs with initial learning rate 0.02 and of 4 per-GPU for 31 epochs with initial learning rate 0.03. All networks are trained with Stochastic Gradient Descent and momentum of 0.9, L2 weight regularization with decay of 0.00017, the learning rate is halved three times, and batch normalization moving averages decay set to 0.9. We use the same hyperparameter values for the $ \lambda^j = 0.1 $ coefficients of the loss as in~\cite{meletis2018heterogeneous}.

\begin{table}
	\setlength\tabcolsep{3.5pt}
	\caption{Overall performance improvements using weak supervision from \textit{OpenScapes} dataset in addition to strong supervision from Cityscapes or Vistas, over the baseline network trained with only per-pixel labels.}
	\label{tab:perf-overall}
	\begin{center}
		\begin{tabular}{ccc|c|c|c|c}
			\multicolumn{3}{c|}{Origin of ground truth} & \multicolumn{4}{c}{results on val splits} \\
			Cityscapes & \multicolumn{2}{c|}{OpenScapes} &\multicolumn{2}{c|}{Cityscapes} & \multicolumn{2}{c}{Vistas} \\
			per-pixel & bound. boxes & image-level & mAcc & mIoU & mAcc & mIoU \\
			\hline
			\ding{51} & & & 77.8 & 68.9 & 53.0 & 43.6 \\
			\ding{51} & \ding{51} & & 79.2 & 70.2 & 52.1 & 43.6 \\
			\ding{51} & \ding{51} & \ding{51} & 79.3 & 70.3 & 52.0 & 43.0
		\end{tabular}
	\end{center}
\end{table}

\begin{table*}
	\caption{Cityscapes per class \textbf{mIoU} (\textbf{\%}) improvements, for the classes, which belong to subclassifiers that receive extra supervision from the weakly labeled \textit{OpenScapes} dataset (100k subsets). Results are grouped per subclassifier.}
	\label{tab:perf-detail-citys}
	\begin{center}
		\begin{tabular}{ccc|cccccc|c||cc|c}
			& & & \multicolumn{7}{c||}{Vehicle subclassifier} & \multicolumn{3}{c}{Human subclassifier} \\
			\cline{4-13}
			& & & {\multirow{5}{*}{\rot{Bicycle}}} & \multirow{5}{*}{\rot{Bus}} & \multirow{5}{*}{\rot{Car}} & \multirow{5}{*}{\rot{Motorc.}} & \multirow{5}{*}{\rot{Train}} & \multirow{5}{*}{ \rot{Truck}} & \multirow{5}{*}{\rot{Overall}} & \multirow{5}{*}{\rot{Person}} & \multirow{5}{*}{\rot{Rider}} & \multirow{5}{*}{\rot{Overall}}\\
			\multicolumn{3}{c|}{Origin of ground truth} & & & & & & & & & & \\
			Cityscapes & \multicolumn{2}{c|}{OpenScapes} & & & & & & & & & & \\
			per-pixel & bound. boxes & image-level & & & & & & & & & & \\
			\hline
			\ding{51} & & & 67.0 & 79.7 & 91.9 & \textbf{52.2} & \textbf{69.3} & 62.3 & 70.4 & 70.2 & 47.9 & 59.0 \\
			\ding{51} & \ding{51} & & 67.8 & \textbf{81.8} & \textbf{92.5} & 50.3 & \textbf{69.3} & 71.4 & \textbf{72.2} & 71.9 & 50.7 & 61.3 \\
			\ding{51} & \ding{51} & \ding{51} & \textbf{67.9} & 79.1 & \textbf{92.5} & 48.7 & \textbf{69.3} & \textbf{75.5} & \textbf{72.2} & \textbf{72.3} & \textbf{51.4} & \textbf{61.9}
		\end{tabular}
	\end{center}
\end{table*}

\section{Experiments}
\begin{figure}
	\centering
	\includegraphics[width=1.0\linewidth]{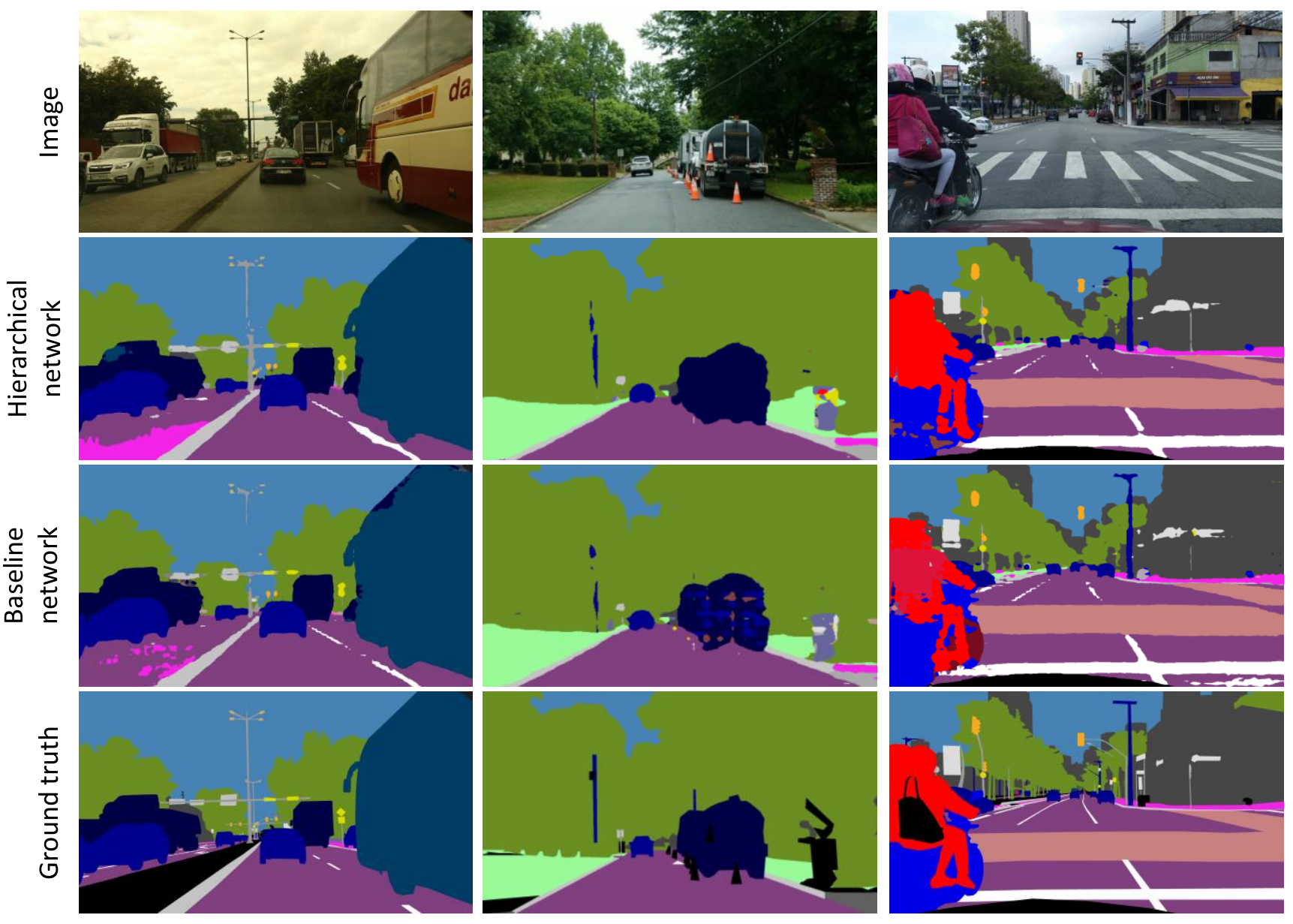}
	\caption{Comparison for Vistas validation split images for the hierarchical network trained on \textit{OpenScapes} bounding boxes subset and Vistas against the baseline network trained on Vistas only. The three classes with biggest improvement in mIoU are Caravan(+11.3\%), On rails(+10.8\%), and Motorcyclist(+9.6\%).}
	\label{fig:vistas-res}
\end{figure}

We evaluate performance using two established multiclass metrics for semantic segmentation~\cite{Cordts2016Cityscapes}, namely mean pixel Accuracy (mAcc) and mean Intersection over Union (mIoU). Metrics for all experiments are evaluated on the validation splits (Cityscapes: 500, Vistas: 2000) of the per-pixel datasets, and are averaged on the last three epochs. In Sec.~\ref{ssec:overall} and~\ref{ssec:perf-per-class} we present overall results and per class results for the classes that receive extra weak supervision. In Sec.~\ref{ssec:perf-effect-of-size} we investigate the effect of the number of examples used from the weakly labeled dataset. Example results from all datasets are shown in Figures~\ref{fig:citys-res} and~\ref{fig:vistas-res}.

\begin{table*}
	\setlength\tabcolsep{3.5pt}
	\caption{Vistas per class \textbf{mIoU} (\textbf{\%}) improvements, for the classes, which belong to subclassifiers that receive extra supervision from the weakly labeled \textit{OpenScapes} dataset (100k subsets). Results are grouped per subclassifier.}
	\label{tab:perfor-detail-vistas}
	\begin{center}
		\begin{tabular}{ccc|ccccccccccc|c||cccc|c}
			& & & \multicolumn{12}{c||}{Vehicle subclassifier} & \multicolumn{5}{c}{Human subclassifier} \\
			\cline{4-20}
			& & & \multirow{6}{*}{\rot{Bicycle}} & \multirow{6}{*}{\rot{Boat}} & \multirow{6}{*}{\rot{Bus}} & \multirow{6}{*}{\rot{Car}} & \multirow{6}{*}{\rot{Caravan}} & \multirow{6}{*}{\rot{Motorcycle}} & \multirow{6}{*}{\rot{On rails}} & \multirow{6}{*}{\rot{Other veh.}} & \multirow{6}{*}{\rot{Trailer}} & \multirow{6}{*}{\rot{Truck}} & \multirow{6}{*}{\rot{Wheeled}} & \multirow{6}{*}{\rot{Overall}} & \multirow{6}{*}{\rot{Person}} & \multirow{6}{*}{\rot{Cyclist}} & \multirow{6}{*}{\rot{Motorcyc.}} & \multirow{6}{*}{\rot{Other rider}} & \multirow{6}{*}{\rot{Overall}} \\
			& & & & & & & & & & & & & & & & & & & \\
			\multicolumn{3}{c|}{Origin of ground truth} & & & & & & & & & & & & & & & & & \\
			Cityscapes & \multicolumn{2}{c|}{OpenScapes} & & & & & & & & & & & & & & & & & \\
			per-pixel & bound. boxes & image-level & & & & & & & & & & & & & & & & & \\
			\hline
			\ding{51} & & & 55.0 & \textbf{26.7} & \textbf{75.0} & \textbf{88.8} & 0.3 & \textbf{54.2} & 38.4 & 16.9 & 0.3 & 65.0 & 7.4 & 38.9 & \textbf{65.5} & \textbf{51.4} & 43.1 & 0.0 & 40.0 \\
			\ding{51} & \ding{51} & & \textbf{56.1} & 21.2 & 73.8 & 88.6 & \textbf{11.6} & 53.9 & \textbf{49.2} & \textbf{18.4} & \textbf{0.9} & \textbf{66.9} & \textbf{10.7} & \textbf{41.0} & 64.7 & 47.1 & \textbf{52.7} & \textbf{0.4} & \textbf{41.2} \\
			\ding{51} & \ding{51} & \ding{51} & 54.5 & 21.2 & 74.0 & 88.4 & 11.4 & 52.8 & 49.0 & 18.1 & 0.8 & 66.0 & 10.6 & 40.6 & 64.6 & 47.1 & 49.9 & 0.3 & 40.5
		\end{tabular}
	\end{center}
\end{table*}

\subsection{Overall results}
\label{ssec:overall}

In Table~\ref{tab:perf-overall} the overall results for Cityscapes~\cite{Cordts2016Cityscapes} and Vistas~\cite{neuhold2017mapillary} are shown. All networks are trained with strong (per-pixel) supervision, from Cityscapes or Vistas, and a combination of weak (per bounding box or image-level or both) supervision from \textit{OpenScapes}. We used the two subsets of \textit{OpenImages} with 100k images each (Sec.~\ref{ssec:openscapes}) and their generated pseudo per-pixel labels, as described in Sec.~\ref{ssec:pp-gt-gen}, mixed in the batch with Cityscapes or Vistas images (see Sec.~\ref{ssec:impl-details} for implementation details).

For Cityscapes, we observe that mAcc and mIoU increase steadily by increasing the amount of weakly labeled data included during training. For Vistas, however, training together with the \textit{OpenScapes} subsets slightly harms the performance. This is possibly due to the diversity of images of Vistas and the large \textit{domain gap} with \textit{OpenScapes}. Overall, we denote that by adding extra supervision for specific classes, mean performance over all classes is not harmed dramatically, and in most cases also boosted.

\subsection{Improvements on classes with weak supervision}
\label{ssec:perf-per-class}
In this Section, we investigate the performance on classes that receive extra weak supervision apart from strong per-pixel supervision. As can be seen in Tables~\ref{tab:perf-detail-citys} and~\ref{tab:perfor-detail-vistas}, overall mIoU of classes belonging to vehicle and human subclassifiers improves in both datasets when adding the \textit{OpenScapes} bounding box labeled subset. Although, in the Cityscapes case, adding the \textit{OpenScapes} image-level labeled subset increases the performance, in the Vistas case it reduces it. We hypothesize that this is due to the \textit{domain gap} between the datasets (see Sec.~\ref{ssec:openscapes-vs-others},~\ref{sec:discussion}). We would also like to mention the big increase for specific classes,~\eg +13.2\% for Cityscapes "Truck" class, +11.3\% for Vistas "Caravan" class, and +10.8\% for Vistas "On rails" class.

\subsection{Effect of weakly labeled dataset size}
\label{ssec:perf-effect-of-size}
In this experiment we train the hierarchical architecture on Cityscapes, together with different portions of the \textit{OpenScapes} bounding boxes labeled subset, with all other hyperparameters fixed, to investigate the effect of the size of the weakly labeled dataset. From Table~\ref{tab:size-matters}, row 2, it becomes clear that without using enough weakly labeled images the performance may even drop. However, when enough weak supervision is provided, row 3 and 4, the performance is enhanced adequately. 

\begin{table}
	\caption{Performance (\%) on Cityscapes with different amount of bounding boxes used to generate pseudo ground truth labels for the weakly labeled dataset.}
	\label{tab:size-matters}
	\begin{center}
		\begin{tabular}{r|c|c}
			per-pixel + \#\textit{images} with bbox GT & mAcc & mIoU \\
			\hline
			$0$ images ($0$ bboxes) & 77.8 & 68.9 \\
			$1k$ images ($17.3k$ bboxes) & 77.4 & 68.4 \\ 
			$10k$ images ($140.4k$ bboxes) & 78.2 & 69.2 \\ 
			$100k$ images ($1185.8k$ bboxes) & 79.2 & 70.2 
		\end{tabular}
	\end{center}
\end{table}

\section{Discussion and future work}
\label{sec:discussion}
The performance of our method heavily depends on two factors: 1) the amount of weak labels and their semantic extent of class connotation, and 2) the \textit{domain gap}~\cite{zhang2017transfer, csurka2017domain, romijnders2019agnostic} between strongly and weakly labeled datasets.

In this work we hypothesized that the images for datasets that are trained simultaneously come from similar domains, and thus features from a common feature extractor can be classified by the same classifier. In reality, this assumption rarely holds, but we leave investigation of this matter and how to solve it during inference to future research. Methods that perform domain agnostic inference, like \cite{romijnders2019agnostic}, can hold solutions for this problem.

Another important matter is the \textit{connotation extent} (the extent of the class name connotation for labeling visually similar objects) of a semantic class. Although in this work, we assumed that classes described by the same high-level semantic concepts, like truck or bus, depict very similar objects across datasets, this is not true in general, and should be investigated in the future. This is visible, for example, in the performance drop for the motorcycle class in Table~\ref{tab:perf-detail-citys}, for which the \textit{connotation extent} for motorcycle objects diverges between Cityscapes and \textit{OpenScapes} datasets.

\section{Conclusion}
We presented a fully convolutional network coupled with a hierarchy of classifiers for simultaneous training on strongly and weakly labeled datasets for semantic segmentation. We collected street scene images from Open Images to generate a weakly labeled dataset called \textit{OpenScapes}. Using \textit{OpenScapes} we showed that the overall performance, as well as the performance for classes that receive extra weak supervision, is increased, provided that enough weak labels are available. Moreover, we examined the effect of the size of the weakly labeled dataset and showed that the performance increase is proportional to the size of the dataset. For our experiments we assumed that the \textit{domain gap} between simultaneously trained datasets is minor, however in other cases it can be a limiting factor, especially when using image-level labels, and should receive attention in future research.


\bibliographystyle{IEEEtran}
\bibliography{biblio}

\end{document}